\documentclass[11pt]{article}

\usepackage{amsmath}
\usepackage{amssymb}
\usepackage{amsthm}
\usepackage{graphicx}
\usepackage{enumerate}
\usepackage{theoremref}
\usepackage{spconf}
\usepackage{bm}
\usepackage{natbib}
\usepackage{xcolor}
\usepackage{caption}
\usepackage{subcaption}
\usepackage{algorithmic}
\usepackage{algorithm}
\usepackage{multirow}
\usepackage{hyperref}
\usepackage{makecell}
\usepackage{epstopdf}

\newtheorem{theorem}{Theorem}[section]

\theoremstyle{definition}

\begin{document}
	
\title{Distributed Randomized Kaczmarz for the adversarial workers\thanks{This work is supported by NSF BIGDATA DMS \#1740325 and NSF DMS \#2011140.}}
\name{Xia Li$^{\star}$ \qquad Longxiu Huang$^{\star}$ \qquad Deanna Needell$^{\star}$}
  
\address{$^{\star}$ University of California, Los Angeles}
\date{}
\maketitle
%
\begin{abstract}
Developing large-scale distributed methods that are robust to the presence of adversarial or corrupted workers is an important part of making such methods practical for real-world problems.
Here, we propose an iterative approach that is adversary-tolerant for least-squares problems. The algorithm utilizes simple statistics to guarantee convergence and is capable of learning the adversarial distributions. Additionally, the efficiency of the proposed method is shown in simulations in the presence of adversaries. The results demonstrate the great capability of such methods to tolerate different levels of adversary rates and to identify the erroneous workers with high accuracy.
\end{abstract}


\section{Introduction}
It is critical for machine learning algorithms and their optimization subroutines to be robust and adversary-tolerant. 
A common approach is to utilize \textit{redundancy}; that is, to request the same computation from multiple workers.
The main challenge with such an approach is to leverage the outputs from these workers  efficiently, and in such a way that even seemingly catastrophic adversarial outputs can be identified and tolerated. 

In this work, our goal is to develop a variation of the randomized Kaczmarz (RK) method \citep{strohmer2009randomized} (see Algorithm \ref{alg:kac_alg}) for adversarial workers to solve the linear system $Ax=b$, where $A \in \mathbb{R}^{m\times n}$, $x\in\mathbb{R}^{n}$ and $b\in \mathbb{R}^m$. 
We assume that there is one central server $w_c$ and $N$ workers in total, among which  $p$ fraction of the workers are adversarial (and unknown). 

\begin{algorithm}[h]
\caption{Randomized Kaczmarz Algorithm}\label{alg:kac_alg}
\begin{algorithmic}
\STATE Select a row index $i_j \in [m]$ with probability $p_{i_j} = \frac{\|A_{i_j}\|_2^2}{\|A\|_F^2}$
    \STATE Update $x_{j+1} = \arg\min_{x\in \mathbb{R}^n}\|x-x_j\|$ 
    \STATE s.t. $A_{i_j}x_{j+1} = b_{i_j}$
    \STATE Repeat until convergence
    \end{algorithmic}
\end{algorithm}

Our approach utilizes simple statistics to identify and ignore adversarial ``errors", and thus the setting in which the adversaries communicate and select among 
$k$ types of errors to output is the most challenging for our approach.
To implement our RK Algorithm, the central server randomly chooses a row index $i$ and broadcasts the data $A_i$ ($i$th row of $A$), $b[i]$ ($i$th entry of $b$) and $x_{j}$ (current estimate) to $n$ randomly chosen workers at each step.
The workers that belong to the $\ell$-th category $C_{\ell}$ take up $p_{\ell}$ fraction of all workers, and $\sum_{\ell=1}^{k}p_{\ell}=p$. 
An adversarial worker $s$ in category $C_{\ell}$ returns $r_{s}=b[i]+e_{\ell}-\langle x_{j},A_i \rangle$ and a reliable worker returns $r_{s}=b[i] -\langle x_{j},A_i \rangle$.
\subsection{Contribution}
Our main contributions are threefold:
(i) develop several efficient algorithms to guarantee accurate estimates for the true solution when adversaries are present, (ii) learn the adversary rate and find the adversarial workers efficiently, (iii) provide theoretical convergence analysis with adversarial workers. 
\subsection{Related work}
\noindent\textbf{Kaczmarz method.~}
The Kaczmarz method is an iterative method for solving linear systems that was first proposed by \citet{karczmarz1937angenaherte} which is also known under the name \textit{Algeberaic Reconstruction Technique} (ART) in computer tomography \citep{gordon1975image,herman1993algebraic,natterer2001mathematics}
and has found various applications ranging from computer tomography to digital signal processing. Later \citet{strohmer2009randomized} propose a randomized version of Kaczmarz, where the the probability of each row being selected is set to be proportional to the Euclidean norm of the row and prove the exponential bound on the
expected rate of convergence. In \citet{Needell2009RandomizedKS}, the author proves that RK converges for inconsistent linear systems to a horizon that depends upon the size of the largest entry of the noise. 

\vspace{4mm}
\noindent\textbf{Distributed computing.~}
In the distributed computation, potential threats are the non-responsive workers (also known as stragglers) and adversarial workers. 
To mitigate the issues with straggling workers
\citep{GORDON1970471,karakus2017straggler} introduce several encoding schemes that embed the redundancy directly in the data itself. 
Later, \citet{bitar2020stochastic} propose an approximate gradient coding scheme for straggler mitigation when the stragglers are random. 
To deal with adversarial workers, \citet{yang2019byrdie} propose a variant of the gradient descent method based on the geometric median in the setting where the workers split all the data. 
\citet{alistarh2018byzantine} discuss the problem of stochastic optimization in an adversarial setting where the workers sample data from a distribution and an $\alpha$ fraction of them may adversarially return any vector. 
These methods either have fundamental statistical barriers to the error any algorithm can achieve or works only when the adversary rate $\alpha$ is less than $\frac{1}{2}$, whereas our algorithm is able to converge to the exact solution even with an adversary rate higher than $\frac{1}{2}$ by utilizing redundancy. 
\section{Algorithm}

\subsection{Algorithms}
In this section, we introduce a simple but efficient \textit{mode}-based algorithm to effectively solve linear systems in the presence of the adversarial workers and identify the potential adversarial workers (which are put in a block-list).
The method detects the mode category based on the returned category size. More specifically, the central worker groups the same results and update the guess with the results from the group with the largest size (i.e., \textbf{mode}).
Given $n$ workers, the expected number of workers from $C_\ell$ is ${np_{\ell}}$\footnote{The central worker $w_c$ uses first $m$ iterations to determine the number of different groups of results during each iteration and take the maximum number.} and number of non-adversarial workers is $n(1-p)$. 
In practice, we randomly choose a group with the maximum group size, its results will be used to update the guess as long as the group size is greater than $n(1-p)$ (see Alg. \ref{alg:block_list} Line \ref{alg:line:7}).  Meanwhile, the block-list is updated through a frequency-based approach throughout the iterations: a counter records if a worker's result fails to be the mode 
during each iteration and after certain number of iterations, a threshold is applied to the counter to identify the potential adversarial workers (see Alg. \ref{alg:block_list} Line \ref{alg:line:9} - \ref{alg:line:11}) Once a worker is in the block-list, it will not be visited again.
We present the full details in Alg.~\ref{alg:block_list}. In the sequel,  the theoretical results are based on  Alg.~\ref{alg:block_list}. 

\begin{algorithm}[h]
\caption{\textsc{Distributed Randomized Kaczmarz with block list}}\label{alg:block_list}
\begin{algorithmic}[1]
    \STATE \textbf{Input}: Initialize \textbf{block-list $B$}, good worker set $D = [N]$, a counter vector $E = 0\in \mathbb{R}^n$, $\text{MaxIter}$, $\text{Tol}$, $c_s = 2 ~\text{Tol}$, checking period $T$.
    \WHILE{$j < \text{MaxIter}$ and $|c_s| > \text{Tol}$, }
    \STATE The central worker $w_c$ selects a row index $i_j \in [m]$ with probability $p_{i_j} = \frac{\|A_{i_j}\|_2^2}{\|A\|_F^2}$
    \STATE Sample $w_1,\cdots,w_n$ uniformly from $D$
    \STATE Broadcast $A_{i_j}$ to $w_1,\cdots,w_n$ 
    \STATE  
    $w_s$ returns $c_s = \frac{\langle A_{i_j},x_i\rangle-b_{i_j}}{\|A_{i_j}\|^2} + e_l$, if  $w_s\in C_l$ 
    \STATE $w_c$ splits  $\{c_s\}_{s = 1}^{n}$ into groups $G_1,\cdots,G_k$ 
   and randomly choose from groups $G_{s}$ that satisfy $|G_{s}| \geq n(1-p)$\label{alg:line:7}
   

   \STATE Update $x^{j+1} = x^{j} + c_{s_{0}}A_{i_j}^\top$
   \STATE Update $E(s) = E(s) + 1,$ if $c_s\notin G_{s_0}$ \footnote{Note for a $n$-dimensional vector $E$, we adopt $E(i)$ to denote the $i$th entry of the vector.}\label{alg:line:9}
   \IF {mod$(j,T) = 0$}\label{alg:line:10}
        \STATE Update $B$ by checking the value of entries in $E$\label{alg:line:11}
        \STATE $D = D\setminus B$
   \ENDIF
   \STATE Update $j=j+1$
    \ENDWHILE
    \STATE \textbf{Output}: $x^j$ and $B$

    \end{algorithmic}
\end{algorithm}
\section{Theoretical Results}
In this section, we study the mode distributions and convergence of Algorithm \ref{alg:block_list} from a theoretical perspective.  For the reader's convenience, we first summarize some important notation in 
  Table \ref{tab:notation}. 
\begin{table}[h]
    \centering
    \footnotesize
    \begin{tabular}{||c|c||}
        \hline
         $N$ & Number of workers in total \\
         \hline
         $n$ & Number of workers used in each iteration\\
         \hline
         $C_{\ell}$ & $\ell$-th error category\\
         \hline
         $p$ & Error ratio\\
         \hline
        $p_{\ell}$ & Error ratio in category $C_\ell$\\
         \hline
        $q$ & \makecell{Probability that there is a mode \\among the outputs of $n$ chosen workers}\\
         \hline
         $q_{\ell}$ & \makecell{probability that there is a mode\\ among the outputs of $n$ chosen workers\\ and the mode is in the category $\ell$} \\
         \hline
         $k$ & Number of adversarial categories\\
         \hline
    \end{tabular}
    \caption{\footnotesize  Notation}
    \label{tab:notation}
    \vspace{-4mm}
\end{table}

\subsection{Adversary Rate Learning}
Alg.~\ref{alg:block_list} utilizes the mode to identify adversaries and achieve convergence. In this section, we will compute the probability that   category $\ell$ is the mode during each iteration.
For simplicity, let $C_0$ denote the category which consists of the ``good" workers   and its fraction is denoted by $p_0 = 1-p$. We set $i_0 = \max( \lceil\frac{n}{k+1}\rceil, \lceil n(1-p)\rceil)$. 
First, we denote $a_{i,\ell}$ as  the coefficients of the term $x^{n-i}$ in the polynomials $\prod\limits_{r=0,r\neq \ell}^{k}q_i^r(x)$ for $\ell = 0,\cdots,k$, where 
\begin{eqnarray*}
q_i^\ell(x) &=&\sum_{j=0}^{i-1}\binom{Np_\ell}{j}x^j.
\end{eqnarray*}
Let $N$ be the number of  total workers, $p_{\ell}$ be the fraction of the category $\ell$, $k$ be the number of  error types. Then the   mode distributions can be   summarized as follows. 
\begin{theorem}
If we choose $n$ workers from $N$ 
workers uniformly
, then  the probability that the mode is in the category $\ell$ is:
\[
 \hat{q}_{mode}^{\ell} =\sum_{i = i_0}^{n}\frac{\binom{Np_\ell}{i}a_{i,\ell}}{\binom{N}{n}}.
 \]
Thus the probability that there is a mode is:
\begin{equation*}
    \begin{aligned}
q =& \sum_{\ell = 0}^{k}\hat{q}_{mode}^{\ell}
=\sum_{i =i_0}^{n}\sum_{l=0}^{k}\frac{\binom{Np_\ell}{i}a_{i,\ell}}{\binom{N}{n}},
\end{aligned}
\end{equation*}
where $\binom{n}{i} = 0$ when $n<i$.
\end{theorem}
{Next, let's consider the probability that a specific worker in category $\ell$ is chosen as a mode worker.}
\begin{theorem}
If we randomly choose $n$ workers from $N$ 
workers uniformly, then 
the probability that the worker $w$ in category $\ell$ is not mode for $s$ times over $S$ iterations is
\[
\binom{S}{s}\left(1-P(w,q_{mode}^{\ell})\right)^s(1-\frac{1}{N}+P(w,q_{mode}^{\ell}))^{(S-s)}
\]
where $P(w,q_{mode}^{\ell})=P(w)P(w \text{ is mode}|w)$ with
\[P(w \text{ is mode}|w) =\sum_{i = i_0-1}^{n-1}\frac{\binom{Np_\ell-1}{i}a_{i,\ell}}{\binom{N-1}{n-1}}, P(w)=\frac{\binom{N-1}{n-1}}{\binom{N}{n}}.\]
\end{theorem}

\subsection{Convergence Guarantee}
\begin{theorem}\label{thm:main}
Let $A\in\mathbb{R}^{d_1\times d_2}$ with $d_1\geq d_2$ and $b,e_1,\cdots,e_{k}\in\mathbb{R}^{d_1}$. Assume that we solve $Ax^*=b$ via Algorithm \ref{alg:block_list}, then 
\begin{equation}
    \begin{aligned}
       \mathbb{E}\|x_{i+1}-x^*\|_2^2
      \leq&\alpha^{i+1}\|x_0-x^*\|_2^2\\
      &+ \frac{1-\alpha^{i+1}}{1-\alpha}\frac{1}{\|A\|_F^2}\sum_{\ell=1}^{k}q_\ell\|e_\ell\|^2,
    \end{aligned}
    \label{eqn:convg}
\end{equation}
where $\sigma_{\min}^2(A)$ is the smallest singular value of $A$, $\alpha=1-\frac{\sigma_{\min}^2(A)}{\|A\|_F^2}$ and $q_{\ell}=\frac{\hat{q}^{l}_{mode}}{q}$.
\\
Additionally, if $\|e_{\ell}\|\leq C$, we thus have \begin{equation}\label{eqn:thm1-2}
    \begin{aligned}
      & \mathbb{E}\|x_{i+1}-x^*\|_2^2\\
      \leq&\alpha^{i+1}\|x_0-x^*\|_2^2+ \frac{1-\alpha^{i+1}}{1-\alpha}\frac{Cq_0}{\|A\|_F^2}.
    \end{aligned}
\end{equation}
\end{theorem}
In the second half of this theorem, we use the fact that $\sum_{\ell=0}^{k}q_\ell=1$. Furthermore, we could assume that $\mathbb{E}\|e_{\ell}\|^2=d\sigma_{\ell}^2$ at each iteration. 

To provide a quantitative understanding of  Theorem \ref{thm:main}, we present several examples in  Tables~\ref{tab:combo_k} and \ref{tab:combo_n} and for simplicity, assume that each error category has the same fraction $p_\ell = p/k$. Thus, all $\hat{q}_{mode}^{\ell}$ are equal. Here $q_0$ is the probability that the algorithm chooses the right mode and $q$ is the probability that there is a mode. Table \ref{tab:combo_k} shows the probability $q$ and $q_\ell$. In these two tables, we
present the values for $\hat{q}_{mode}^{\ell},\hat{q}_{mode}^{0},q$ and $q_0$ by varying the number of error types $k$, the number of chosen workers $n$ and the adversarial rate $p$. These two tables are generated by solving a linear system with a row-normalized matrix $A\in\mathbb{R}^{1000\times 100}$.
 As $k$ increases, $q_\ell$ decreases and $q_0$ increases. Therefore, the error bound in equation \eqref{eqn:thm1-2} 
decreases with respect to $k$ and thus reaches better convergence results. When $k$ is large enough, $ q_{\ell}\approx0$. Therefore, when the noise is random error and there is a mode for the step-size, the mode will be the correct mode.
 As $n$ increases, there is a similar decrease effect and therefore a better convergence.
\begin{table}[h]
    \centering
    \footnotesize
    \scalebox{0.85}{
    \begin{tabular}{||c|c|c|c|c|c||}
         \hline
         $p$ &$k$ & $\hat{q}_{mode}^{\ell}$ & $\hat{q}_{mode}^{0} $  & $q$  &$q_0$\\
         \hline
         \multirow{3}{4em}{\quad $0.8$}
         &$5$ &$0.1$ &$0.16$ &$0.67$  &$0.15$\\
         &$10$ & $0.04$ &$0.21$ &$0.57$  &$0.36$\\
        &$15$ &$0.02$ & $0.23$ & $0.48$  &$0.46$\\
         \hline
        \multirow{3}{4em}{\quad$0.2$} 
        &$3$ &$0.002$ &$0.63$ &$0.64$  &$0.98$\\
        &$5$ &$8\times10^{-4}$ &$0.65$ &$0.65$  &$0.99$\\
         &$10$ &$2\times10^{-4}$ &$0.66$ &$0.67$  &$0.99$\\
         &$15$ &$2\times10^{-4}$ &$0.685$ &$0.689$ &$0.99$\\
         \hline
    \end{tabular}
    }
    \caption{\footnotesize  Total number of workers $N=100$, number of chosen workers $n=5$.}
    \label{tab:combo_k}
\end{table}
\begin{table}[h]
    \centering
    \footnotesize
    \scalebox{0.85}{
    \begin{tabular}{||c|c|c|c|c|c||}
         \hline
         $p$ &$n$ & $\hat{q}_{mode}^{\ell} $ & $\hat{q}_{mode}^{0}$  & $q$ &$q_0$\\
         \hline
         \multirow{3}{4em}{\quad $0.8$}
         & 10 &0.099 &0.18 &0.67 &0.26\\
         &15 & 0.099 &0.2 &0.7  &0.29\\
         & 20 &0.097 &0.23 &0.71  &0.31\\
         \hline
        \multirow{3}{4em}{\quad$0.2$} 
        &10 &$7\times10^{-6}$ &0.904 &0.90 &$1 - 5\times10^{-6}$ \\
        &15 &$5\times10^{-7}$ &0.97 &0.97 &$1- 3\times10^{-6}$\\
        &20 &$1\times10^{-7}$ &0.99  &0.99 &$1- 6\times10^{-7}$\\
         \hline
    \end{tabular}
    }
    \caption{\footnotesize  Total number of workers $N=100$, number of error categories $k=5$.}
    \label{tab:combo_n}
\end{table}

\section{Simulation}
\begin{figure}[!h]
    \centering
    \begin{subfigure}[b]{0.235\textwidth}
        \centering
         \includegraphics[width=\textwidth]{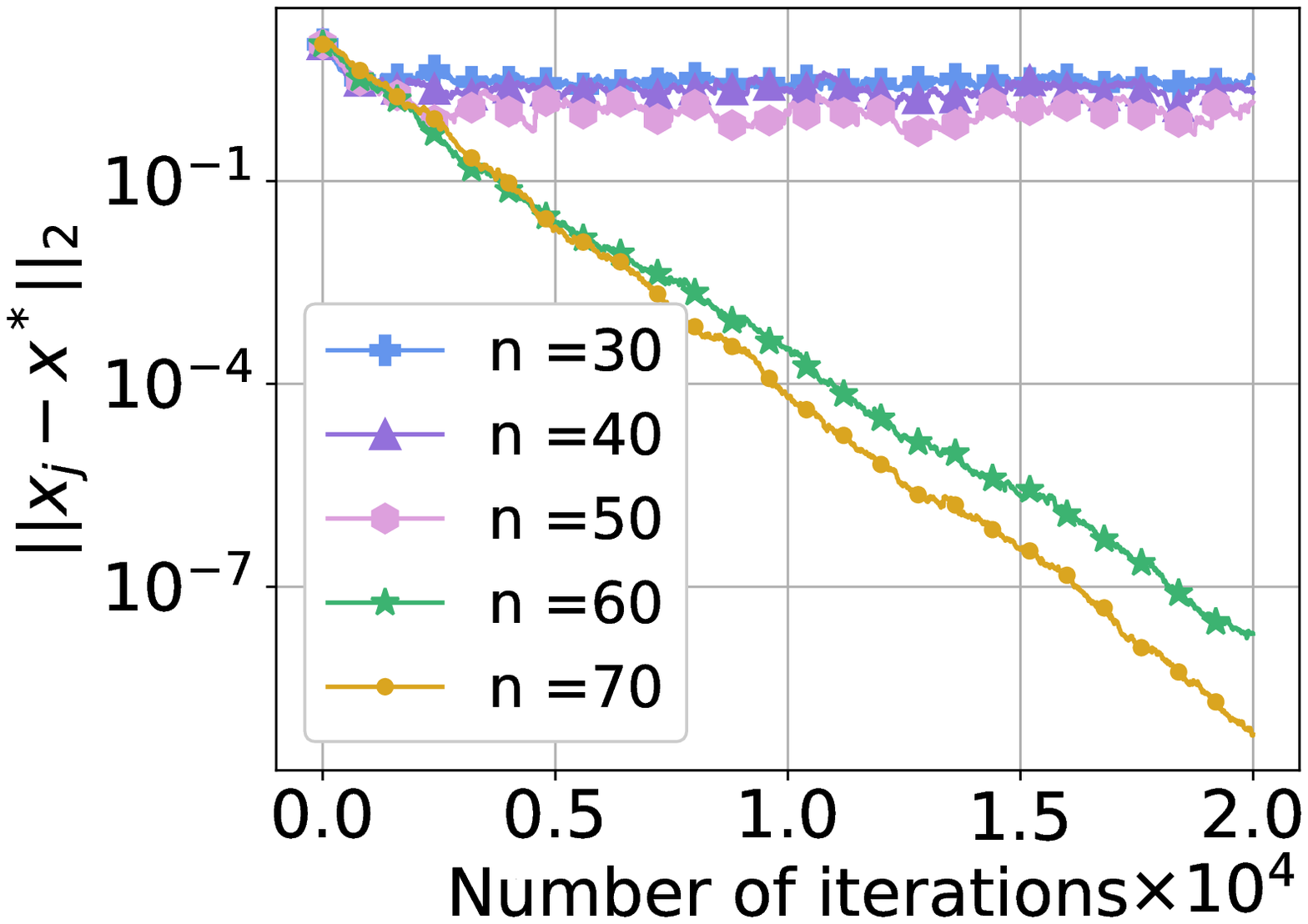}
         \caption{$p = 0.8$, without block-list}
         \label{fig:p = 0.8, wo block-list}
    \end{subfigure}
    \begin{subfigure}[b]{0.235\textwidth}
    \centering
     \includegraphics[width=\textwidth]{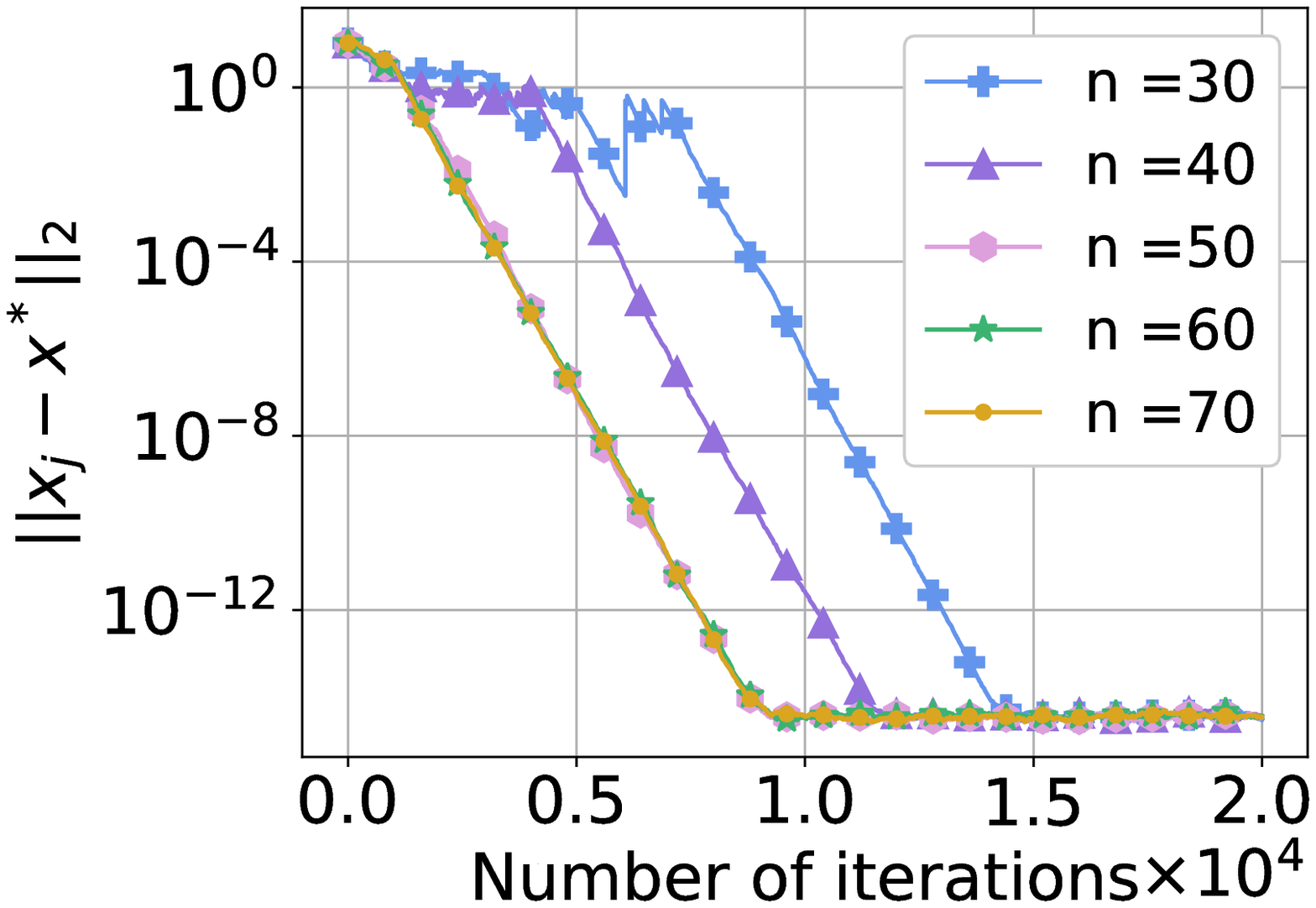}
     \caption{$p = 0.8$, with block-list}
     \label{fig:p = 0.8, with block-list}
    \end{subfigure}
    \\
    \begin{subfigure}[b]{0.235\textwidth}
        \centering
         \includegraphics[width=\textwidth]{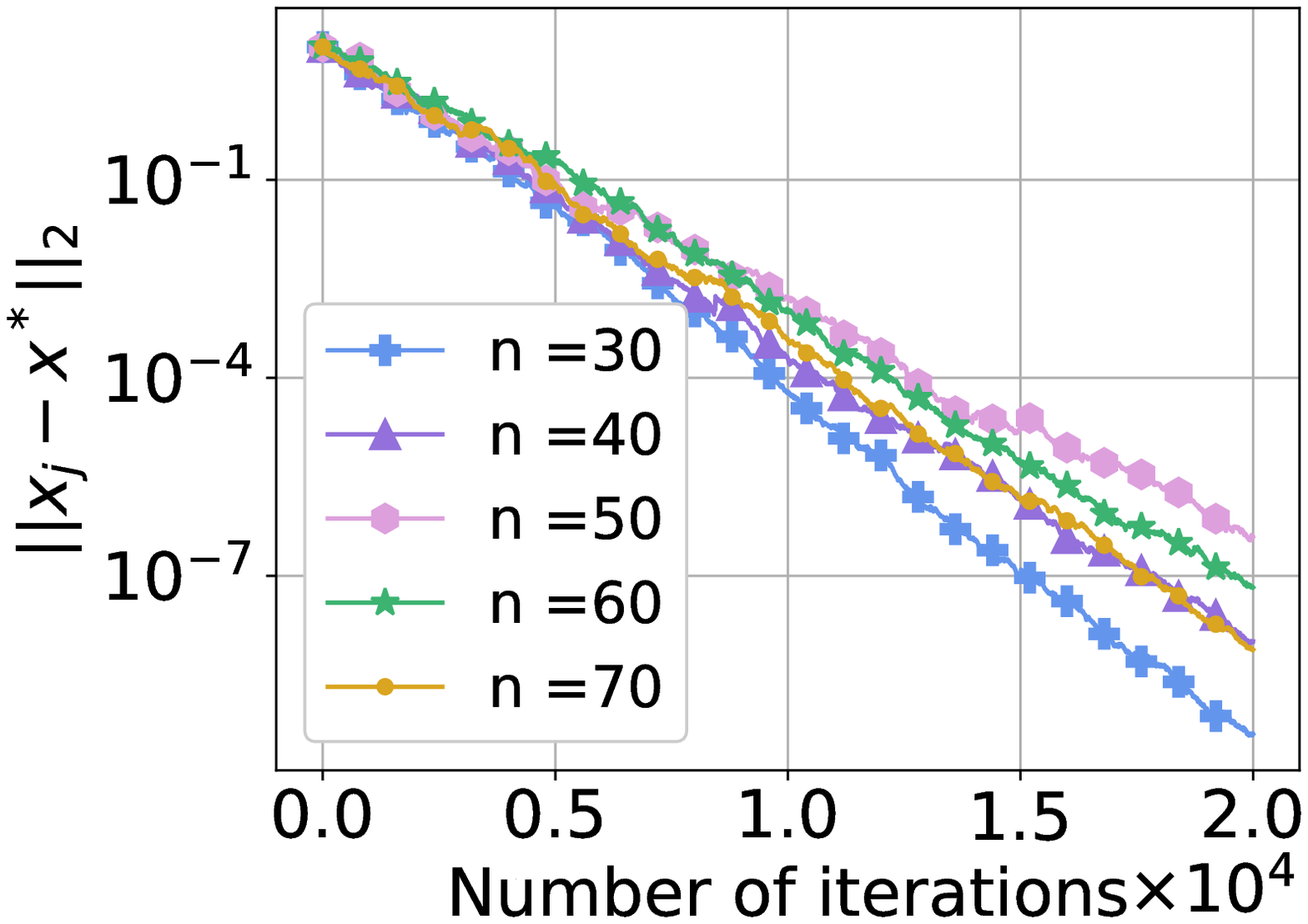}
         \caption{$p = 0.2$, without block-list}
         \label{fig:p = 0.2, wo block-list}
    \end{subfigure}
    \begin{subfigure}[b]{0.235\textwidth}
        \centering
         \includegraphics[width=\textwidth]{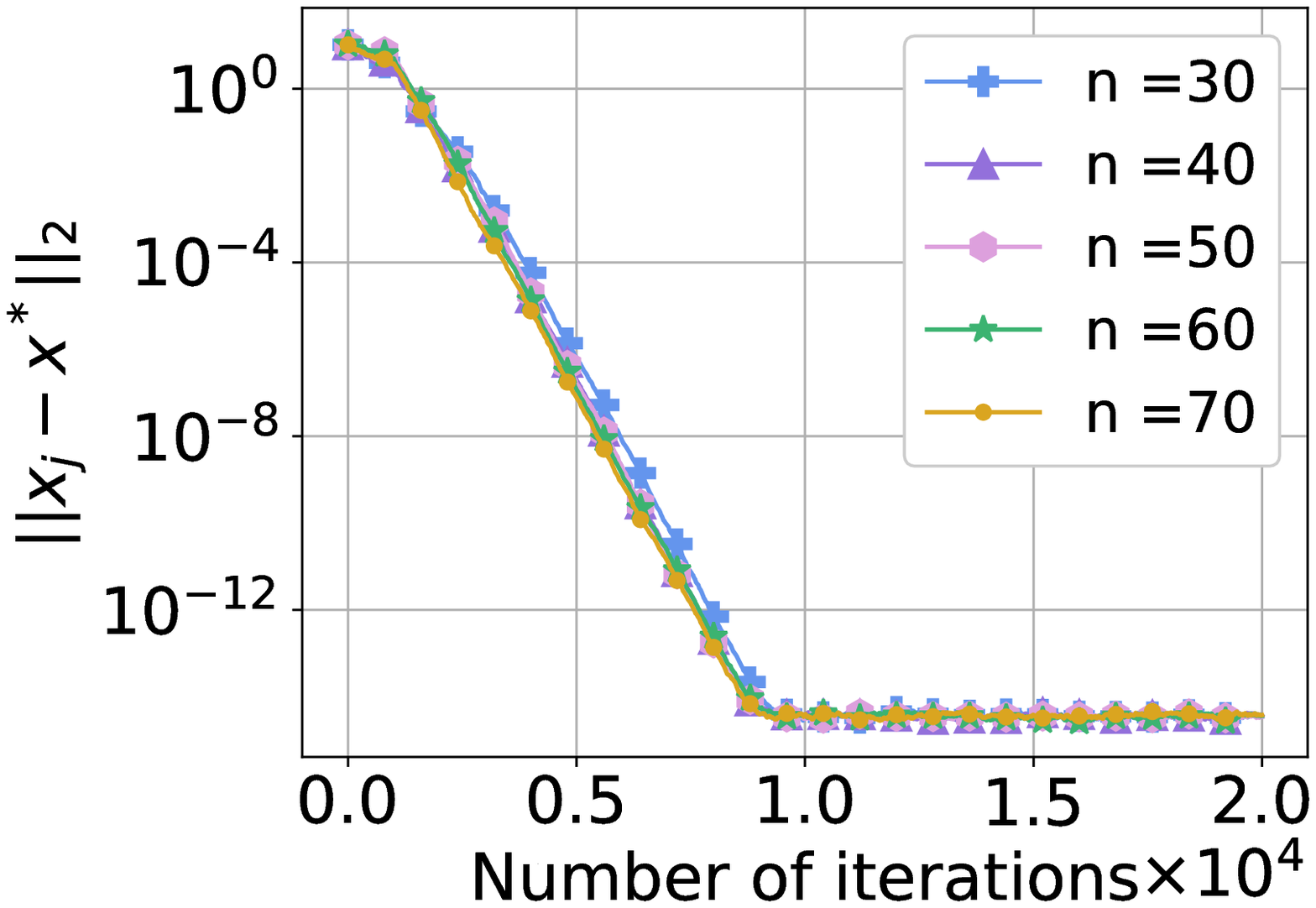}
         \caption{$p = 0.2$, with block-list}
         \label{fig:p = 0.2, with block-list}
    \end{subfigure}
    \caption{\footnotesize  The effect of the number of the chosen workers:
    $10$ error categories  and choosing $n$ workers  from $100$ workers with  $n= 30,40,50,60,70$.}
\label{fig:repeated}
\end{figure}
In this section, we test the performance of our approach for solving consistent/inconsistent linear systems. The simulation shows how the number of the chosen workers, the adversary rate and the number of the error categories affect the performance.  
\begin{figure}[!h]
    \centering
    \begin{subfigure}[b]{0.235\textwidth}
        \centering
         \includegraphics[width=\textwidth]{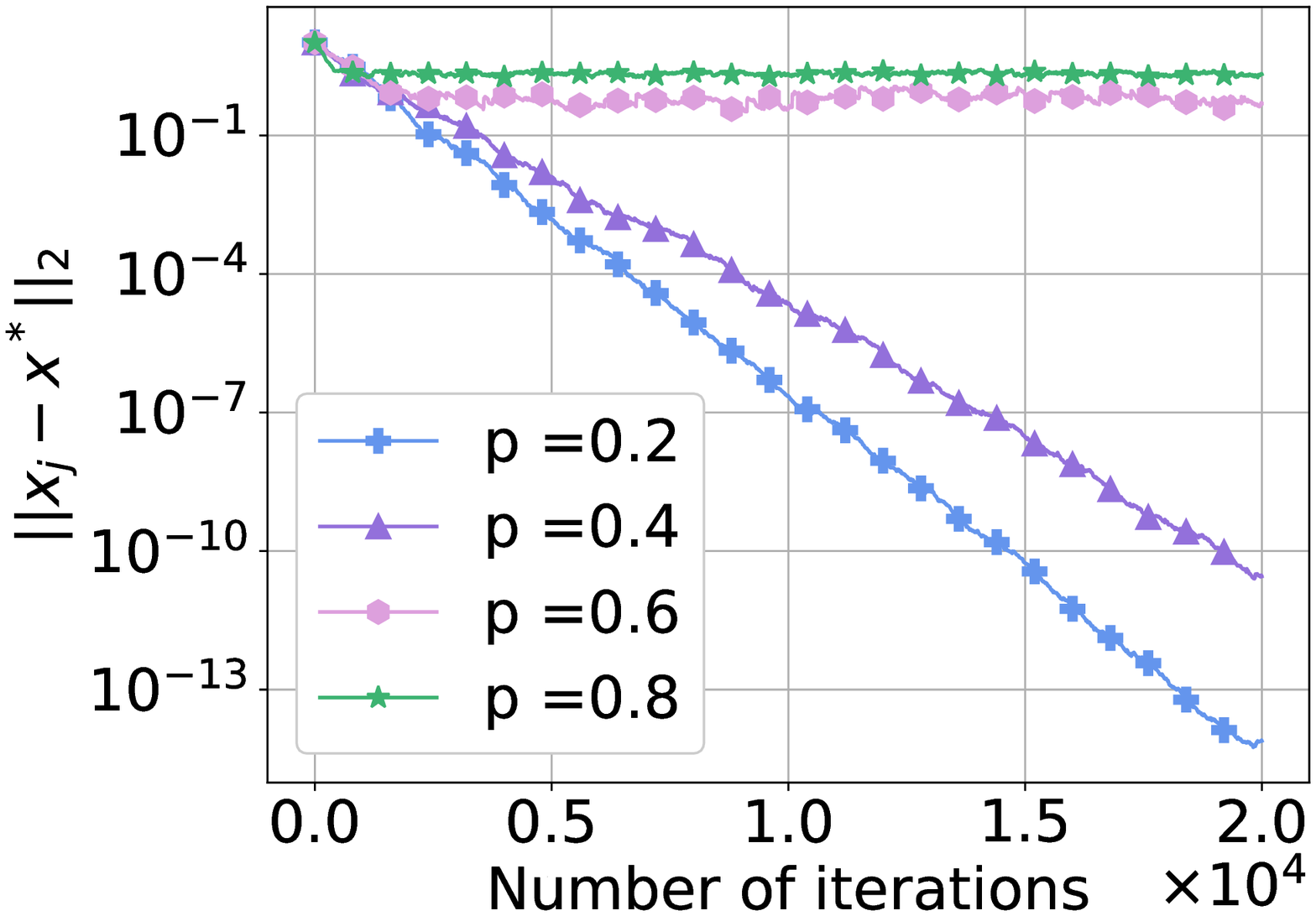}
         \caption{Without block-list}
         \label{fig:diifp, wo block-list}
    \end{subfigure}    
    \begin{subfigure}[b]{0.235\textwidth}
        \centering
         \includegraphics[width=\textwidth]{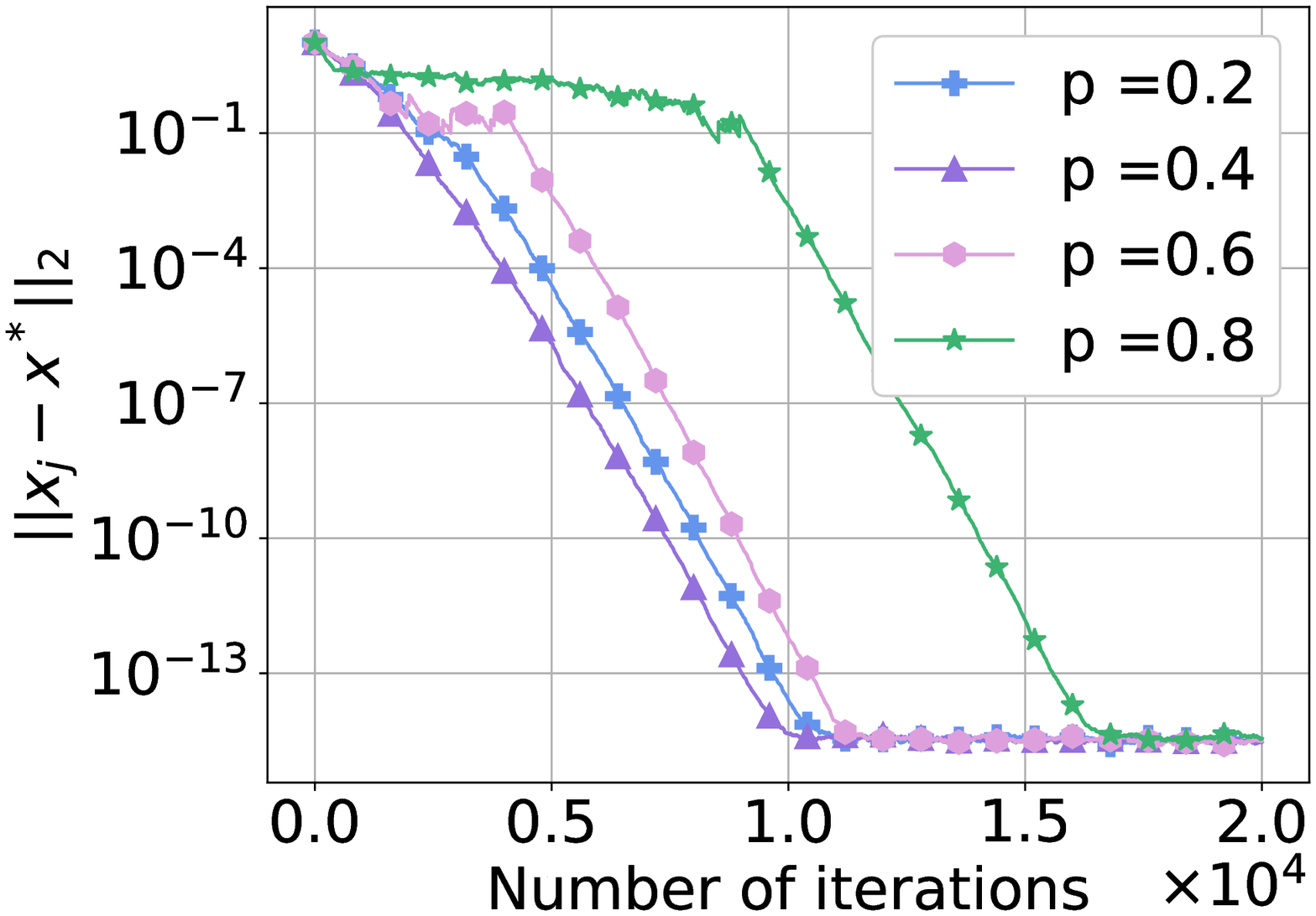}
         \caption{With block-list}
         \label{fig:diffp, w block-list}
    \end{subfigure}
    \caption{\footnotesize  The effect of the adversary rate:
     $10$ error categories  and choosing $10$ workers from $100$ workers.}
    \label{fig:diffp}
\end{figure}

\begin{figure}[!h]
    \centering
      \begin{subfigure}[b]{0.235\textwidth}
        \centering
         \includegraphics[width=\textwidth]{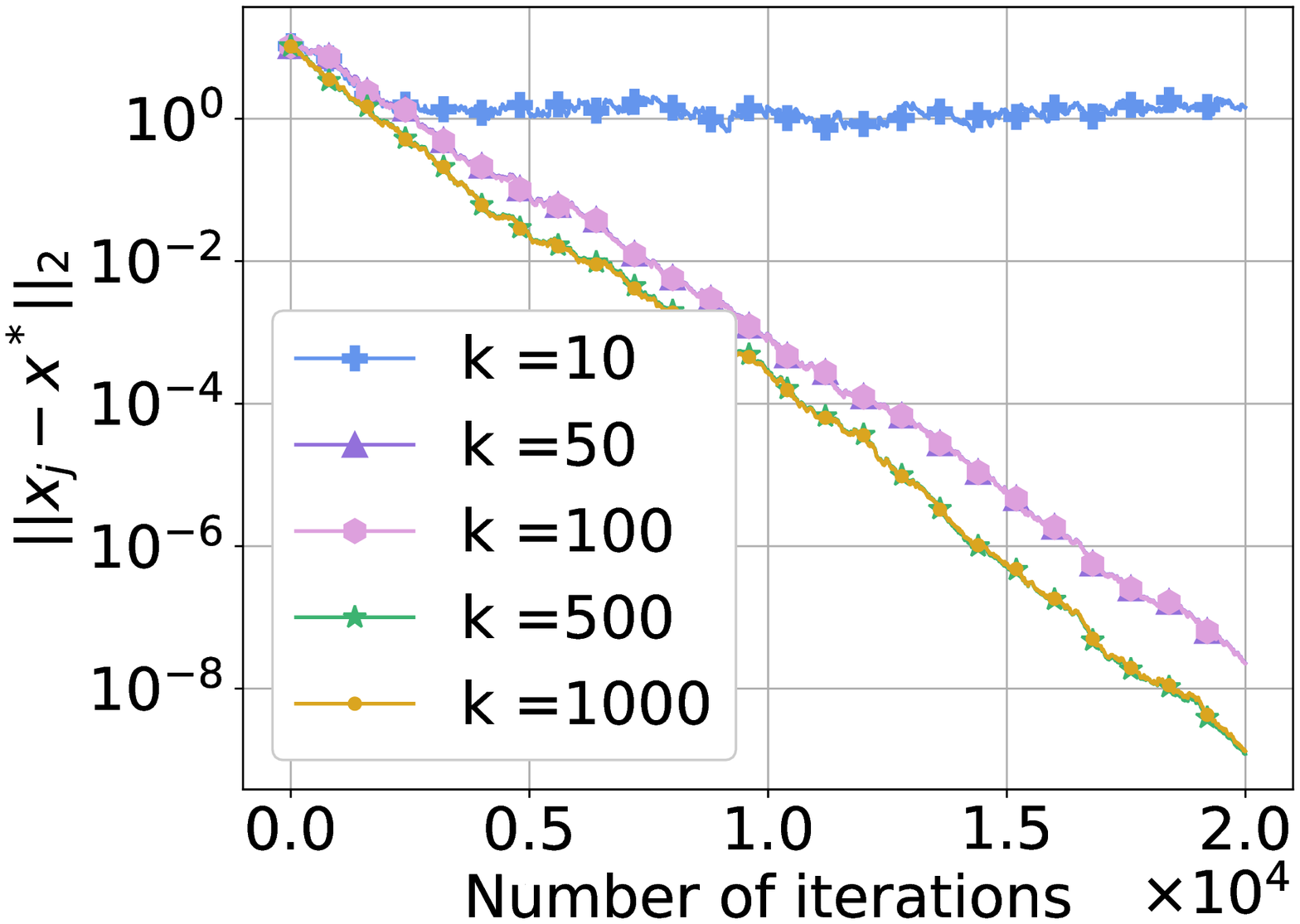}
         \caption{Without block-list}
         \label{fig:diffkp = 0.8,wo block-list}
    \end{subfigure}
    \begin{subfigure}[b]{0.235\textwidth}
        \centering
         \includegraphics[width=\textwidth]{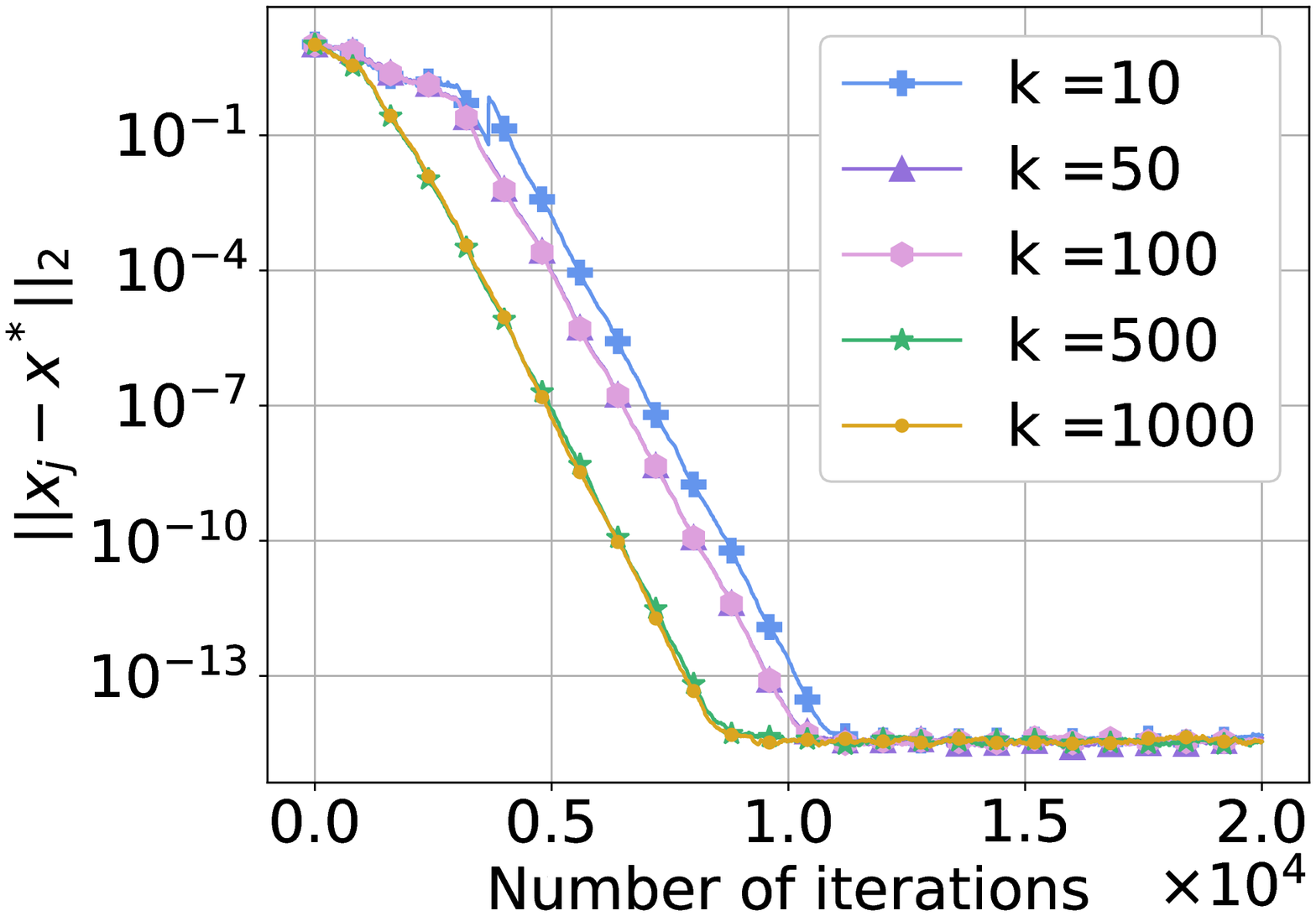}
         \caption{ With block-list}
         \label{fig:diffkp = 0.8,w block-list}
    \end{subfigure}
    \caption{\footnotesize  The effect of the error category number: 
    the adversary rate $p=0.8$ and choosing $10$ workers   from $100$.}
    \label{fig:diffk}
    \vspace{-4mm}
\end{figure}

\begin{figure}[!h]
    \centering
       \begin{subfigure}[b]{0.235\textwidth}
        \centering
         \includegraphics[width=\textwidth]{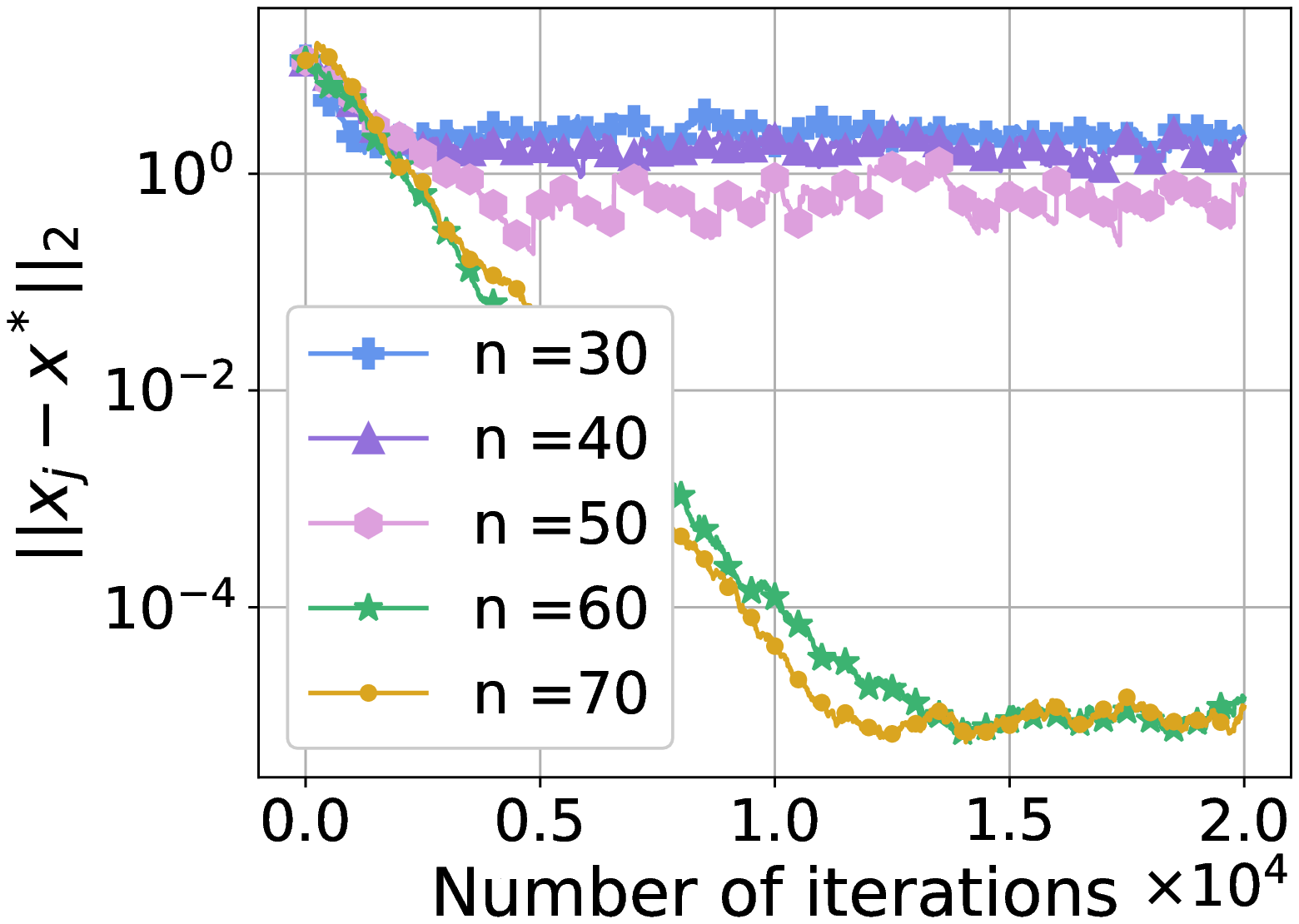}
         \caption{Without block-list}
         \label{fig:noise_p = 0.8,wo block-list}
    \end{subfigure}
    \begin{subfigure}[b]{0.235\textwidth}
        \centering
         \includegraphics[width=\textwidth]{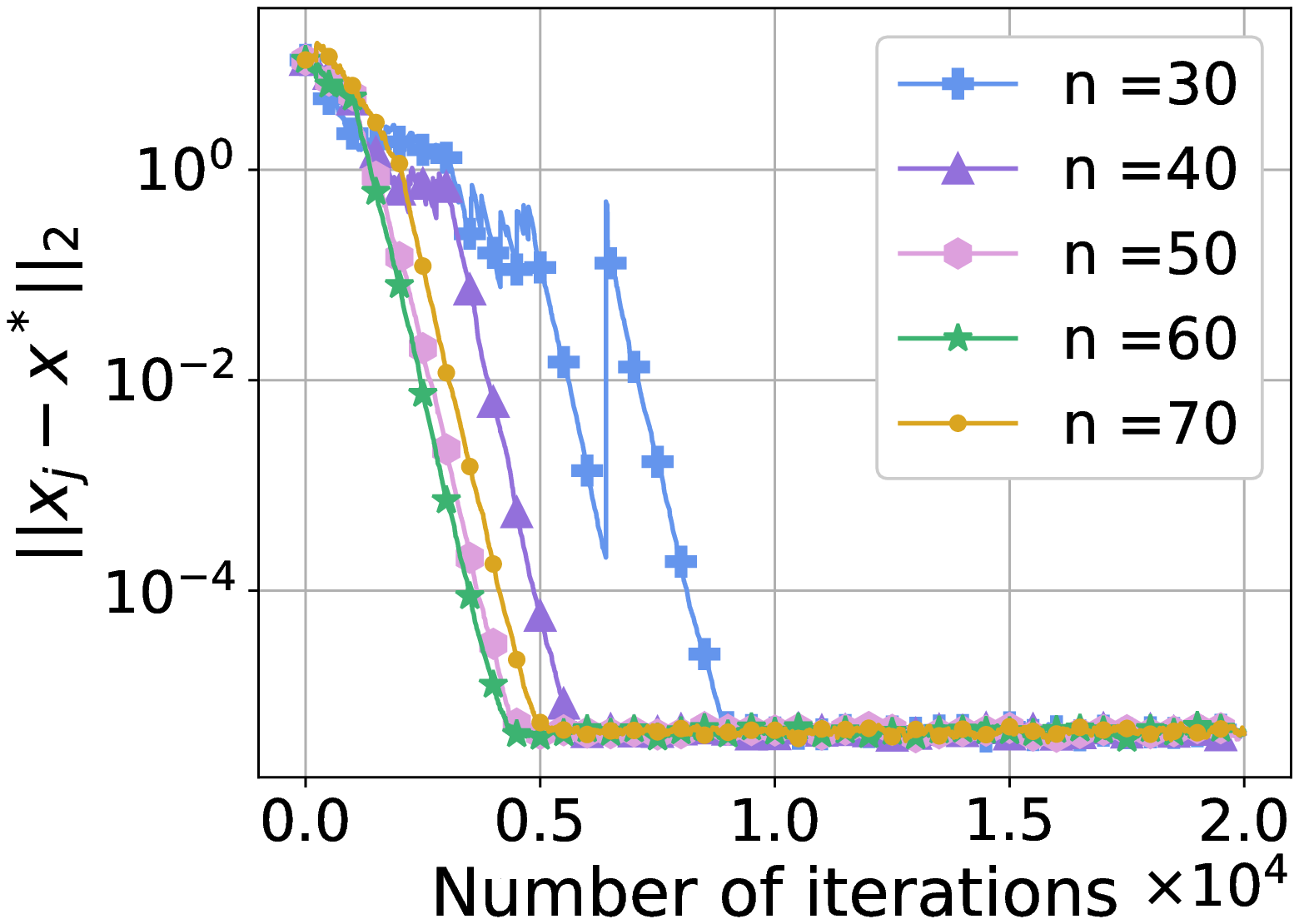}
         \caption{With block-list}
         \label{fig:noise_p = 0.8,w block-list}
    \end{subfigure}
\caption{\footnotesize The effect of the number of chosen workers $n$ for inconsistent linear systems: $Ax=b+e$ with random noise $e \sim \text{uniform}(10^{-4})$, adversary rate $p = 0.8$, $10$ error categories, and choosing $n$ workers  from $100$  with  $n= 30,40,50,60,70$.}
    \label{fig:noise_10-2b}
     \vspace{-3mm}
\end{figure}

In the simulation, we randomly generate a row-normalized matrix $A\in \mathbb{R}^{m\times n}$, $x\in \mathbb{R}^n$ and set $b = Ax$.  $Ax = b$ is solved via Alg.~\ref{alg:block_list} with/without  block-list. At each iteration, one row of $A$ is randomly chosen and distributed to all chosen workers. 
  Fig.~\ref{fig:repeated} presents the effect of the   number of the chosen workers $n$ and shows the convergence results for the mode detection method without/with the block-list, respectively. When the adversary rate $p$ is $20\%$ and $80\%$  over $100$ workers. Without using the block-list method, the central server would use a corrupted step size to update and oscillate around the solution, as shown in Fig.~\ref{fig:p = 0.8, wo block-list}. When $p = 20\%$, the error is as low as $10^{-10}$, although the result is not as good as the one with the block-list, the trade-off is the extra storage for the block-list. As the number of chosen workers  $n$ increase from $30$ to $70$ the convergence is faster for both without or with the block-list. The algorithm shows good performance for solving inconsistent linear systems (by adding some random noise  to $b$) as well (see Fig.~\ref{fig:noise_10-2b}). Fig.~\ref{fig:diffp} presents the effect of the adversary rate. As the adversary rate $p$ increases, the accuracy  decreases. Even though the adversary rate is large,  the final results are still satisfying by solving the system via the algorithm with a block-list.  In additional, Fig.~\ref{fig:diffk} shows the effect of  the number of category types $k$. The algorithm converges as $k \rightarrow \infty$, i.e. with random noises. With the block-list, the convergence error is less than $10^{-13}$ after a sufficient number of iterations. 
  Furthermore, we summarize performance using precision and recall scores in Table \ref{tab:pnr}. With adequate workers, the algorithm is able to identify the adversaries with an accuracy over $98\%$ and demonstrates the effectiveness of block-list method.
\begin{table}[h]
    \centering
    \footnotesize
    \begin{tabular}{||c|c|c|c|c|c||}
    \hline
        $n$ & $30$ & $40$ &$50$ &$60$ &$70$\\
        \hline
         Precision &$0.988$ &$0.988$ &$1$ &$0.988$ &$1$ \\
         \hline
         Recall &$0.988$ &$0.988$ &$1$ &$0.988$ &$1$\\
         \hline
    \end{tabular}
    \caption{\footnotesize  Precision and recall of the block-list method when number of used workers $n = 30, 40, 50, 60, 70$ and adversary rate $p=80\% $.}
    \label{tab:pnr}
\end{table}

 \vspace{-5mm}
\section{Conclusion}
It is of great significant for optimization algorithms to be robust and resistant to adversary. 
We propose efficient algorithms and provide theoretical convergence guarantee in the presence of the adversarial workers. Our algorithm is able to deal with different adversarial rates even when $p > 0.5$, and at the same time, identify the adversarial source. We also present the effect of several important parameters of the adversaries and of the anti-adversaries strategy, namely, the number of error categories $k$, the adversary rate $p$, and the number of chosen workers $n$ at each iteration.\\

\bibliography{refer.bib}

\begin{thebibliography}{11}
\providecommand{\natexlab}[1]{#1}
\providecommand{\url}[1]{\texttt{#1}}
\expandafter\ifx\csname urlstyle\endcsname\relax
  \providecommand{\doi}[1]{doi: #1}\else
  \providecommand{\doi}{doi: \begingroup \urlstyle{rm}\Url}\fi

\bibitem[Alistarh et~al.(2018)Alistarh, Allen-Zhu, and
  Li]{alistarh2018byzantine}
Dan Alistarh, Zeyuan Allen-Zhu, and Jerry Li.
\newblock Byzantine stochastic gradient descent.
\newblock \emph{arXiv preprint arXiv:1803.08917}, 2018.

\bibitem[Bitar et~al.(2020)Bitar, Wootters, and
  El~Rouayheb]{bitar2020stochastic}
Rawad Bitar, Mary Wootters, and Salim El~Rouayheb.
\newblock Stochastic gradient coding for straggler mitigation in distributed
  learning.
\newblock \emph{IEEE Journal on Selected Areas in Information Theory}, 2020.

\bibitem[Gordon et~al.(1970)Gordon, Bender, and Herman]{GORDON1970471}
Richard Gordon, Robert Bender, and Gabor~T. Herman.
\newblock Algebraic reconstruction techniques (art) for three-dimensional
  electron microscopy and x-ray photography.
\newblock \emph{Journal of Theoretical Biology}, 29\penalty0 (3):\penalty0
  471--481, 1970.

\bibitem[Gordon et~al.(1975)Gordon, Herman, and Johnson]{gordon1975image}
Richard Gordon, Gabor~T Herman, and Steven~A Johnson.
\newblock Image reconstruction from projections.
\newblock \emph{Scientific American}, 233\penalty0 (4):\penalty0 56--71, 1975.

\bibitem[Herman and Meyer(1993)]{herman1993algebraic}
Gabor~T Herman and Lorraine~B Meyer.
\newblock Algebraic reconstruction techniques can be made computationally
  efficient (positron emission tomography application).
\newblock \emph{IEEE transactions on medical imaging}, 12\penalty0
  (3):\penalty0 600--609, 1993.

\bibitem[Kaczmarz(1937)]{karczmarz1937angenaherte}
S~Kaczmarz.
\newblock Angenaherte auflosung von systemen linearer glei-chungen.
\newblock \emph{Bull. Int. Acad. Pol. Sic. Let., Cl. Sci. Math. Nat.}, pages
  355--357, 1937.

\bibitem[Karakus et~al.(2017)Karakus, Sun, Diggavi, and
  Yin]{karakus2017straggler}
Can Karakus, Yifan Sun, Suhas Diggavi, and Wotao Yin.
\newblock Straggler mitigation in distributed optimization through data
  encoding.
\newblock In \emph{Advances in Neural Information Processing Systems}, pages
  5434--5442, 2017.

\bibitem[Natterer(2001)]{natterer2001mathematics}
Frank Natterer.
\newblock \emph{The mathematics of computerized tomography}.
\newblock SIAM, 2001.

\bibitem[Needell(2009)]{Needell2009RandomizedKS}
Deanna Needell.
\newblock Randomized {K}aczmarz solver for noisy linear systems.
\newblock \emph{BIT Numerical Mathematics}, 50:\penalty0 395--403, 2009.

\bibitem[Strohmer and Vershynin(2009)]{strohmer2009randomized}
Thomas Strohmer and Roman Vershynin.
\newblock A randomized kaczmarz algorithm with exponential convergence.
\newblock \emph{Journal of Fourier Analysis and Applications}, 15\penalty0
  (2):\penalty0 262, 2009.

\bibitem[Yang and Bajwa(2019)]{yang2019byrdie}
Zhixiong Yang and Waheed~U Bajwa.
\newblock Byrdie: Byzantine-resilient distributed coordinate descent for
  decentralized learning.
\newblock \emph{IEEE Transactions on Signal and Information Processing over
  Networks}, 5\penalty0 (4):\penalty0 611--627, 2019.

\end{thebibliography}
\bibliographystyle{plainnat}

\end{document}